# Acoustical Quality Assessment of the Classroom Environment

Marian George
Dept. of Comp. and Sys. Eng.
Faculty of Engineering
Alexandria University, Egypt
marian.george@alexu.edu.eg

Moustafa Youssef
Dept. of Comp. Sc. and Eng.
Egypt-Japan Univ. of Sc. and Tec. (E-JUST)
Alexandria, Egypt
moustafa.youssef@ejust.edu.eg

*Abstract*— **Teaching is one of the most important factors affecting any education system. Many research efforts have been conducted to facilitate the presentation modes used by instructors in classrooms as well as provide means for students to review lectures through web browsers. Other studies have been made to provide acoustical design recommendations for classrooms like room size and reverberation times. However, using acoustical features of classrooms as a way to provide education systems with feedback about the learning process was not thoroughly investigated in any of these studies. We propose a system that extracts different sound features of students and instructors, and then uses machine learning techniques to evaluate the acoustical quality of any learning environment. We infer conclusions about the students' satisfaction with the quality of lectures. Using classifiers instead of surveys and other subjective ways of measures can facilitate and speed such experiments which enables us to perform them continuously. We believe our system enables education systems to continuously review and improve their teaching strategies and acoustical quality of classrooms.**

## I. Introduction

A high-quality education system is essential to any university. Providing better equipment and resources for the students, or increasing flexibility in coursework requirements that must be met are among the attempts seeking better learning atmosphere. However, teaching remains a major, if not the major, factor in assessing the quality of any learning environment. In classrooms, the primary methods of communication are speech and listening. There are different factors affecting the effectiveness of the learning conditions like speech intelligibility in rooms, reverberation times and background noise levels. Non-optimal classroom acoustical conditions directly affects speech perception by students and instructors and, thus, reduces student learning efficiency. This specially affects students who have a hearing loss or are working in a second language. We find a lot of studies within this field [3]-[5] trying to achieve better environment for students to learn in. While the teaching environment is important for students, it is equally important to teachers. Teachers suffer from health problems related to their voice. These problems are strongly related to the working environment of teachers including the acoustics of the classrooms, the behavior of students and overall working conditions. For the teacher, in the long run, this voice load due to speaking in the classroom can result in voice disorders such as hoarseness, voice fatigue and can even force teachers to retire early from their profession. This can be a serious problem for governments and private schools [6].

Speech processing is the study of speech signals and processing methods of these signals. Information like gender, speaker identity, dialect, emotion, and other characteristics can be extracted from speech signals using speech processing techniques. Speech processing along with information extraction techniques has been used in many contexts like natural language processing [1], diagnosis and screening of vocal and voice diseases [2]. However, very few research efforts have been conducted in the classroom context.

Having continuous analysis of acoustical quality in classrooms is very beneficial for improving learning environments. Several studies have been made to provide acoustical design recommendations for classrooms like room size, reverberation times and amount of acceptable background noise. Nevertheless, these design decisions do not cancel the need for an on-going assessment of the acoustical quality in such classrooms to capture different relationships between the instructor's sound features like his/her sound power level, the students' sound features like the amount of background noise they create and physical classroom features. These relationships can be automatically inferred through extracting the required features and using machine learning techniques and algorithms to classify them. Using classifiers instead of surveys and other subjective ways of measures can facilitate and speed such experiments which enables us to perform them continuously. Currently, computers are involved in teaching mostly through interactive learning technologies trying to achieve better teacher-student interaction. In this paper, we show that the role of technology in teaching can be broadened to assist in judgment of lectures quality, making decisions regarding effective teaching strategies, and increasing students' satisfaction.

## II. Related Work

Many research efforts were reported which try to create an intelligent classroom; one which understands what the lecturer wants to do and responds accordingly [7]-[9]. McConnell Engineering 13 (MC13) classroom [7] provides a mechanism for the capture, collation, and synchronization of digital notes, written on an electronic whiteboard or digital tablet, with an audio-visual recording of the lecturer. The system then allows students to review lectures, either in randomly accessed portions, or in their entirety, through a conventional web browser, either from networked university computers or home computers connected by modem. Northwestern University's Intelligent Classroom project [9] produces an automated video of a lecture; the camera tracks the lecturer's focus, infers the speaker's intentions and then the presentation cameras pan, tilt and zoom to best capture what is important at every moment in the lecture. Students can then review lecture videos later, making up for what they have missed.

These systems and others [10][11][12] aim at facilitating the professors' use of a variety of presentation modes in class, enabling students to review lecture content and materials, and enabling alternative means of student evaluation to be implemented. None of them addressed the teaching strategy itself or different characteristics of the learning environment like the lecturer speech level or acoustics of the classroom.

Other researchers attempted to study different voice parameters of instructors and acoustics of the lecture rooms. In [13]-[15], different groups of speakers where studied where the voice parameters of speakers with and without voice problems were measured. The parameters measured were sound pressure level at a distance of 1m from the speaker and pitch. The changes in these parameters were studied to analyze voice changes during work for teachers. We can also find studies focusing on speech intelligibility in classrooms and advisable background noise level [16][17]. In [18], J. Brunskog et al. investigate whether objectively measurable parameters of the rooms can be related to an increase of the voice sound power produced by speakers and to the speaker's subjective judgments about the rooms. Reverberation time and other physical attributes, the sound power level were measured. They found that there is a correlation between acoustic properties of a room and the sound power level of the lecturer. It was concluded that these changes mainly have to do with the size of the room and to the gain produced by the room. As we can see, the correlation between different parameters of the instructor and the students was not thoroughly investigated in any of these studies. The main focus was to study the teacher's satisfaction with the classroom conditions and changes in the instructor's voice parameters throughout the working day.

In recent years, many approaches have been developed to address the problem of robust speech analysis, using feature-normalization algorithms, microphone arrays, and other approaches. Accordingly, speech processing techniques can be used in noisy environments yielding useful and acceptable results. In this paper, we use speech features like pitch, and A-weighted sound level combined with machine learning techniques as GMM and K-NN classifiers to infer conclusions about students' satisfaction with the quality of lectures. We determine different features of students' noise like intensity of noise and location of noise, and correlate these features with instructor features as sound level and/or gender. Our system enables education systems to continuously review and improve their teaching strategies.

The data our system relies on is simply short-interval recordings, around 10-20 seconds each, taken throughout the duration of a lecture. After preprocessing of the data, we extract the required features and feed them to a k-NN classifier to automatically classify the lecture as noisy or quiet. The same recordings are used to extract instructor features such as pitch and sound level; these features are then fed to a GMM classifier which automatically identifies the gender of the speaker. Different relations between instructor features and noisy/quiet lectures are then inferred. By comparing the A-weighted sound level from multiple recorders placed in different positions inside a classroom, we are able to determine the approximate location of the source of noise. An instructor can use such information, for example, through interacting with students in the noisy locations frequently to draw their attention and limit their side talks.

Our contribution can be summarized in:
1) Using technology in classrooms in a different way other than interactive learning tools
2) Using speech processing techniques in a new context to measure lecturing quality and propose strategies for better teaching environments
3) Running experiments in real environments, without affecting the usual normal activity of teachers or students or imposing any burden on them
4) Analyzing the results and inferring factors affecting lecture quality and relations among these factors

In section 3, we talk about the overall architecture of our system. In section 4, we explore the details of the different component of the system. Experiments and performance evaluation of the system are shown in section 5.

III. SYSTEM ARCHITECTURE

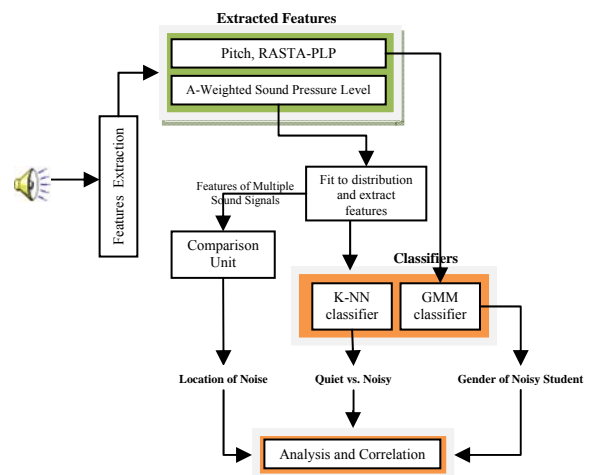

Fig. 1. System architecture: features are extracted from sound recordings, fed into classifiers, results are analyzed.

Figure 1 presents the overall architecture of our system. We describe the high level processing flows here and describe the details in the next section. We begin by capturing sound clips, each clip lasts for a short duration of time, around 10 seconds, throughout a lecture. Sound clips are then pre-processed for features extraction. The goal of preprocessing is to smooth the extracted contours through using overlapped frames and to reduce ripples in the speech spectrum by multiplying each frame by a Hamming window. Pre-processing also involves omitting silent intervals from analysis in gender classification. Features are then extracted from the processed time samples. We extract A-weighted sound level for determination of the state of the lecture as being quiet or noisy. Sound level is also used to determine the approximate location of noise as we will explain. Pitch and RASTA-PLP features are extracted for gender classification following the approach in [20]. Pitch features are further filtered by a medium filter to increase the robustness of pitch. Average sound power level of each sound clip is extracted and stored to be used in teacher-student differentiation as explained in the following paragraph.

Having extracted the required features for gender classification, Gaussian mixture densities are used to model the feature vectors comprised of the pitch and the 10th order RASTA-PLP coefficients. Considering the features extracted from voiced speech frames only, two different Gaussian mixture densities are trained for male and female speech, pmale(u|μ,Σ,α) and pfemale(u| μ,Σ,α). Gender classification is then performed using normalized log-likelihood. Given an utterance from an unknown speaker, the normalized log-likelihood of LLmale(x) and LLfemale(x) are calculated. If LLmale(x) > LLfemale(x), the speaker is determined as a "male", otherwise, the speaker is determined as a "female". Assuming that the instructor is the main speaker in most sound clips taken throughout a single lecture and that the instructor's sound level varies within a very small range throughout the duration of the lecture, our system concludes that the instructor's sound power level will be the one corresponding to the majority of the stored sound power levels of all sound clips of the lecture. The sound power level of a single clip is then compared against the pre-determined instructor sound power level in the teacher-student differentiation unit to determine whether the speaker is the teacher or a student. Using this information along with the gender of the speaker identified earlier, we can now automatically determine the gender of the instructor or any speaking student.

For noise estimation, a histogram of the extracted A-weighted sound pressure level frequency distribution is then fit to a normal distribution curve. We tried fitting non-symmetrical curves like Weibull distribution curves; we observed that the results were not improved over -symmetrical- normal-distribution curves. The mean and standard deviation of the curve were recorded. The features are then used to train a K-NN classifier to classify a sound file as noisy or quiet. Using k-fold cross validation with k = 3, error rate = 0.1 is achieved using 5-NN classifier.

After classifying each sound file as noisy or quiet, the lecture as a whole is classified as noisy if the number of noisy recordings throughout the lecture exceeds the number of quiet recordings; conversely, the lecture is classified as quiet if the number of quiet recordings exceeds the number of noisy recordings.

To estimate the approximate location of noise, we fit normal distribution curves to the A-weighted sound pressure levels of four processed recordings at different positions in the classroom (front left, front right, back left, back right). The four recordings are recorded simultaneously. The maximum average level among the four curves indicates the position of the noise, whether it s the front left of the classroom, the front right, the back left or the back right.

The analysis and classification results reached above are then used to investigate the correlation between different attributes of a lecture. Possible relationships between different characteristics like the amount of students' noise, most common location of noise, instructor's speech level, instructor's gender and other characteristics are investigated and correlation coefficients are computed. Results of this stage are very helpful in improving the lecture's environment, as well as making decisions regarding effective teaching strategies, and increasing students' satisfaction.

IV. SYSTEM DESIGN

A. Features Extraction

*1) Sound Pressure Level:* In atmospheric sounding and noisy environments, room noise level is the background sound pressure level at a given location. The sound pressure level Lp in decibels (dB) is defined by

$$L_p = 20\lg(P/P_0) \qquad (1)$$

Where $P_0$ is a reference pressure (20E-6 Pa), which corresponds to the threshold of audibility.

As presented in [19], the A-weighted sound level is (used extensively in the evaluation of the noise. The A-weighting filter was designed to mimic the response of the human hearing system.

*2) Pitch and RASTA-PLP:* The major difference between male and female speech is pitch. In general, female speech has higher pitch range than male speech. Therefore, pitch is used, usually combined with other features, in gender identification. Many pitch determination algorithms (PDAs) have been proposed [21]. In our system, we used the autocorrelation model [22] which is one of the most popular PDAs for its simplicity and explanatory power.

As there is some overlap in the pitch ranges of male and female speech, other features are combined with pitch to increase gender classification accuracy. In perceptual linear prediction (PLP), the knowledge about the human hearing system is used to process the recorded waveform in such a way that only the perceptually relevant details remain. Relative spectral (RASTA) method is proved to be more robust for recognition in noisy environments as it compensates for the channel effects in recognizers. Following the approach of [20], the relative spectral perceptual linear predictive (RASTA-PLP) coefficients are combined with pitch to be used for gender classification.

B. Classification

*1) Students Noise Estimation:* Sound-pressure-level frequency distributions (in the statistical sense indicating the

proportion of time for which the level took given values, not the acoustical sense) were determined and plotted. These distributions were then fit to a normal distribution curve which gave similar results to –non-symmetrical- Weibull-distribution curves. The mean and standard deviation of the curve were recorded. The features are then used to train a K-NN classifier to classify a lecture as noisy or quiet. Figure 2 shows the means and standard deviation of about 80 recordings. Using k-fold cross validation with k = 3, error rate = 0.1 is achieved using 5-NN classifier. The lecture as a whole is classified as noisy if the number of noisy recordings throughout the lecture exceeds the number of quiet recordings; conversely, the lecture is classified as quiet if the number of quiet recordings exceeds the number of noisy recordings.

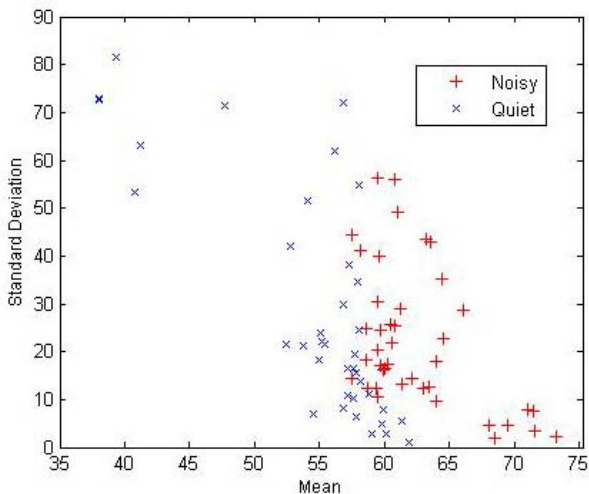

Figure 2. Mean and standard deviation of sound pressure level of noisy and quiet lectures.

*2) Teacher/Student Gender Identification:* A Gaussian Mixture Model (GMM) is a parametric probability density function represented as a weighted sum of Gaussian component densities representing a parametric probability density function. GMM parameters are estimated from training data using the iterative Expectation-Maximization (EM) algorithm or Maximum A Posteriori (MAP) estimation from a well-trained prior model. The EM equations for training a GMM can be found in [25].

Gaussian mixture densities are used to model the 12-dimensional feature vectors comprised of the pitch and the 10th order RASTA-PLP coefficients. We only consider the features that are extracted from voiced speech frames. Two different Gaussian mixture densities for male and female speech are trained, $p_{male}(x, \lambda)$ and $p_{female}(x, \lambda)$.

Gender classification is then performed using normalized log-likelihood. Given an utterance from an unknown speaker, the normalized log-likelihood of $LL_{male}(x)$ and $LL_{female}(x)$ are calculated. If $LL_{male}(x) > LL_{female}(x)$, the speaker is determined as a "male", otherwise, the speaker is determined as a "female".

*C. Analysis and Correlation*

Possible relationships between different characteristics like the amount of students' noise, most common location of noise, instructor's speech level, instructor gender and other characteristics are investigated.

*1) Noise Location Estimation:* Sound pressure level varies as the distance from the speaker is changed. This property is used in determining the approximate location of noise in a classroom. For estimating the quality of a teaching environment, highly accurate speech localization is not necessary, therefore we approximate the location of noise to whether it is at front left, front right, back left or back right of the classroom. We fit normal distribution curves to the A-weighted sound pressure levels of four processed recordings at different positions in the classroom. The four recordings are recorded simultaneously. The maximum average level among the four curves indicates the position of the noise, whether it s the front left of the classroom, the front right, the back left or the back right. In practice, this can be used to determine the most common location of noise.

*2) Classification of Lectures as Noisy or Quiet with Instructor's Speech Level:* We compare lecture noise level (categorized as low, high) and instructor's speech level (categorized as low, medium high), using chi-square test of independence. If association was observed between the two variables, we, then, measure the strength of association by comparing proportions. The differences in proportions must range between -1 and +1. A difference close to one in magnitude indicates a high level of association, while a difference close to zero represents very little association. For visual examination of the results, we fit a normal distribution curve to the frequency distribution of instructors' speech levels for quiet lectures. Similarly, we fit another curve to the frequency distribution of instructors' speech levels for noisy lectures. We then compare the mean value of both curves and the overlap between the curves.

*3) Classification of Lectures as Noisy or Quiet with Instructor's Gender:* Similarly, we compare lecture noise level and instructor's gender using chi-square test of independence and differences in proportions.

## V. EVALUATION

*A. Data Collection*

Our experiments were performed in two university classrooms located in the second floor of a 5-storey building with about eight classrooms in each floor. Outside noise is comprised mainly of student sounds and rustling of tree leaves but generally, no significant noise from outside the classroom occurred. Typical classroom noises, such as coughing and chair movement, were not removed. The classroom has dimension of 12m by 10m, and includes seven rows of student chairs, each row having 10 chairs. Both male and female instructors, teaching a wide variety of subjects were involved. In most cases the classes consisted of lectures, with the instructor being the main source of speech. About three classes were tutorial sessions, with students discussing different exercises with the instructor. Recordings were made using voice recorders of Nokia N97 mobile phones.

Our experiments for student noise estimation and student/teacher gender identification were performed in real settings. Recordings were taken through thirty live lectures; each lecture has duration of one hour. Data were collected for 10-20 sec, in around twenty blocks throughout the whole class in different positions.

Experiments for location of noise estimation were performed in controlled settings where we placed four microphones: two microphones in the first row of the classroom and another two in the seventh row of the classroom. Each two microphones in the same row were placed one on the far left and the other on the far right of the classroom. Two students were seated nearest to each microphone and were recorded talking for around 5 minutes. Results from the four microphones were compared for approximate determination of location of talking students.

### B. Performance

*1) Classification of Lectures as Noisy or Quiet:* For each lecture, k-NN classifier was used to classify each sound recording as noisy or loud. Then, the lecture as a whole was classified as noisy if the number of noisy recordings exceeded the number of quiet recordings; conversely, the lecture was classified as quiet if the number of quiet recordings exceeded the number of noisy recordings. We achieved a classification accuracy of 94.2% when using k = 5. Figure 3 shows classification accuracy of thirty lectures when using different values of k for the k-NN classifier.

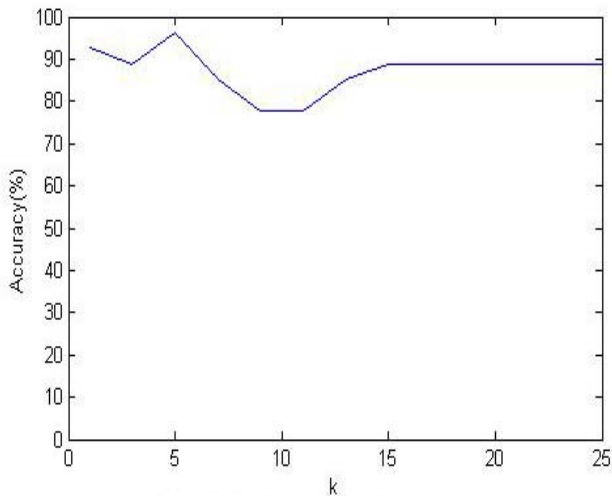

Fig. 3. Lectures classification accuracy with different values of k for the k-nn classifier.

### C. Correlation Results

*1) Noise Location Estimation:* There was a significant difference between the sound pressure levels measured near the group of speakers and those measured in other corners of the classroom. Accordingly, 100% accuracy was achieved in determining the approximate location of the noise: front right, front left, back right, or back left. Figure 5 shows the frequency distribution curves of sound pressure levels of recordings taken near the speakers and far from them as fit by normal-distribution curves; there is no overlap between the two curves.

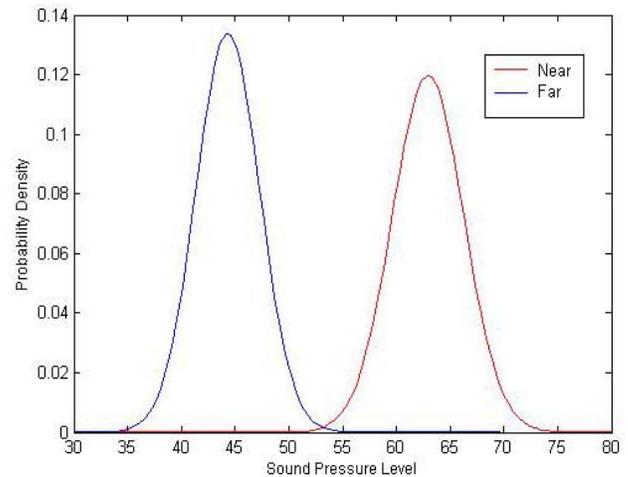

Fig. 4. Frequency distribution curves of sound pressure levels of recordings taken near the speakers and far from them.

*2) Classification of Lectures as Noisy or Quiet with Instructor's Speech Level:* Comparing lecture noise level (categorized as low, high) and instructor's speech level (categorized as low, medium high) using chi-square test of independence yielded a P-value of 0.0073. This P-value indicated that there is some association between both variables. By comparing proportions, we got a difference in proportions of 0.75 which showed a high level of association between both variables. Figure 6 shows the frequency distribution curves of instructors' speech levels as fit by normal distribution curves. It is shown that the speech level mean value for quiet lectures is relatively higher than the speech level mean value for noisy lectures. An interesting observation is that the speech level mean values overlap for quiet and noisy lectures; this is due to when the instructor's voice is relatively high, students pay attention and the lecture is generally quiet. On the other hand, when the students are noisy, the instructor sometimes has to raise his/her voice to overcome the background noise.

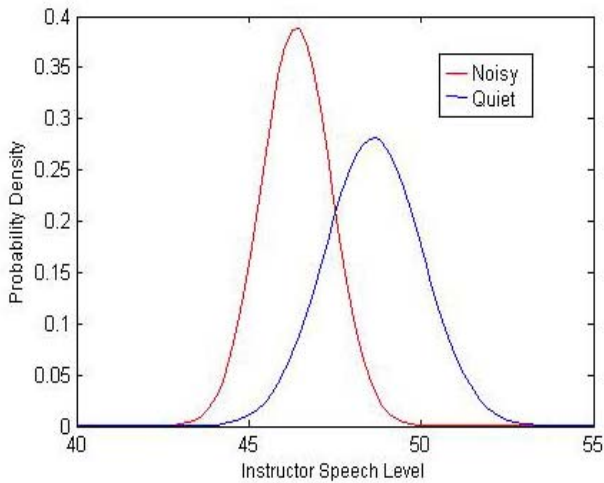

Fig. 5. Frequency distribution curves of instructor's speech levels as fit by normal distribution curves.

*3) Classification of Lectures as Noisy or Quiet with Instructor's Gender:* Comparing lecture noise level (categorized as low, high) and instructor's gender (categorized as male, female) using chi-square test of independence yielded a P-value of 0.0574. This P-value indicated that there is some association between both variables. By comparing proportions, we got a difference in proportions of 0.4 which showed a low level of association between both variables. So, in conclusion lecture noise level is weakly related to the instructor's gender. This generally agrees to previous studies concerning the evaluation of instructors based on students ratings [26][27]. In studies that analyze large samples of courses from a variety of disciplines, the consistent result is that there are no significant differences in ratings due to systematic gender bias. Female and male faculty do not appear to be rated higher or lower by students by virtue of their gender.

## VI. CONCLUSION AND FUTURE WORK

In this paper, we propose a system to automatically and continuously evaluate the acoustical quality of the classroom environment. By extracting different sound features of the students and the instructor then using machine learning techniques, we produce classification and correlation results useful in providing recommendations and guidelines. We evaluate the performance of the different components of our system as follows:

- Classification of lectures as noisy or quiet: using k-NN classifier, we achieved a classification accuracy of 94.2% when using k=5.
- Noise location estimation: there was a significant difference between the sound pressure levels measured near the group of speakers and those measured in other corners of the classroom. Accordingly, 100% accuracy was achieved in determining the approximate location of the noise: front right, front left, back right, or back left.
- Correlation between the amount of noise generated by students and the instructor's speech level: we use chi-square test of independence then we compare proportions to measure the level of association between both variables. Our experiments showed a high level of association between both variables.
- Correlation between the amount of noise generated by students and the instructor's gender: we use chi-square test of independence then we compare proportions to measure the level of association between both variables. Our experiments showed that lecture noise level is weakly related to the instructor's gender. This generally agrees to previous studies concerning the evaluation of instructors based on students ratings [26][27].

We evaluate our system in four university classrooms at the beginning of the next semester, February 2012. The continuous acoustical assessment of each lecture will guide the instructor to take measures and actions in order to improve the quality of the learning environment. For example, by automatically determining the location of students who continually produce noise, the instructor can pay additional attention to this corner, or try to communicate with them more. Also, the statistical correlation results will provides guidelines like advisable instructor speech levels appropriate in different classrooms.

In future research, we plan to expand the features we extract to include other non-acoustical characteristics of the instructor. We will also work on producing other correlation results to further enhance the learning environment.